\documentclass[lettersize,journal]{IEEEtran}
\usepackage{amsmath,amsfonts}
\usepackage{algorithm}
\usepackage{array}
\usepackage[caption=false,font=normalsize,labelfont=sf,textfont=sf]{subfig}
\usepackage{textcomp}
\usepackage{stfloats}
\usepackage{url}
\usepackage{verbatim}
\usepackage{graphicx}
\usepackage{cite}
\usepackage{soul}
\usepackage{amssymb}
\usepackage{latexsym}
\usepackage{url}
\usepackage{xcolor, soul, color}
\usepackage{lipsum}  
\usepackage{newtxmath, mleftright}
\usepackage{stackengine}
\usepackage{bm}
\usepackage{multirow}
\usepackage{algpseudocode}

\usepackage{balance}
\usepackage{graphicx} 
\usepackage{float}
\usepackage{siunitx}
\usepackage{gensymb}
\usepackage{diagbox}
\usepackage{url}
 
\begin{document}

\title{Augmented Reality and Human-Robot Collaboration Framework for Percutaneous Nephrolithotomy}

\author{Junling Fu$^{\dagger}$,~\IEEEmembership{Student Member,~IEEE,} Matteo Pecorella, Elisa Iovene,~\IEEEmembership{Student Member,~IEEE,} \\Maria Chiara Palumbo,~\IEEEmembership{Student Member,~IEEE}, Alberto Rota,~\IEEEmembership{Student Member,~IEEE}, \\Alberto Redaelli, Giancarlo Ferrigno,~\IEEEmembership{Senior Member,~IEEE},  and Elena De Momi,~\IEEEmembership{Senior Member,~IEEE}
\thanks{
Junling Fu, Matteo Pecorella, Elisa Iovene, Maria Chiara Palumbo, Alberto Rota, Alberto Redaelli, Giancarlo Ferrigno, and Elena De Momi are with the Department of Electronics, Information and Bioengineering, Politecnico di Milano, 20133, Milan, Italy. 
($\dagger$, \textit{Corresponding author}, junling.fu@polimi.it)
         }
}

\markboth{}%
{Shell \MakeLowercase{\textit{et al.}}: A Sample Article Using IEEEtran.cls for IEEE Journals}


\maketitle

\begin{abstract}
During Percutaneous Nephrolithotomy (PCNL) operations, the surgeon is required to define the incision point on the patient's back, align the needle to a pre-planned path, and perform puncture operations afterward. The procedure is currently performed manually using ultrasound or fluoroscopy imaging for needle orientation, which, however, implies limited accuracy and low reproducibility. This work incorporates Augmented Reality (AR) visualization with an optical see-through head-mounted display (OST-HMD) and Human-Robot Collaboration (HRC) framework to empower the surgeon’s task completion performance. In detail, Eye-to-Hand calibration, system registration, and hologram model registration are performed to realize visual guidance. A Cartesian impedance controller is used to guide the operator during the needle puncture task execution. Experiments are conducted to verify the system performance compared with conventional manual puncture procedures and a 2D monitor-based visualization interface. The results showed that the proposed framework achieves the lowest median and standard deviation error of $3.17\pm1.36$ mm in translation and $1.2\pm0.9 \degree$ in orientation across all the experimental groups, respectively. Furthermore, the NASA-TLX user evaluation results indicate that the proposed framework requires the lowest workload score of $42.5\pm13.7$ for task completion compared to other experimental setups. The proposed framework exhibits significant potential for clinical application in the PCNL task, as it enhances the surgeon’s perception capability, facilitates collision-free needle insertion path planning, and minimizes errors in task completion.

\end{abstract}


\section{Introduction}
\begin{figure}
    \centering
    \includegraphics[scale=0.4]{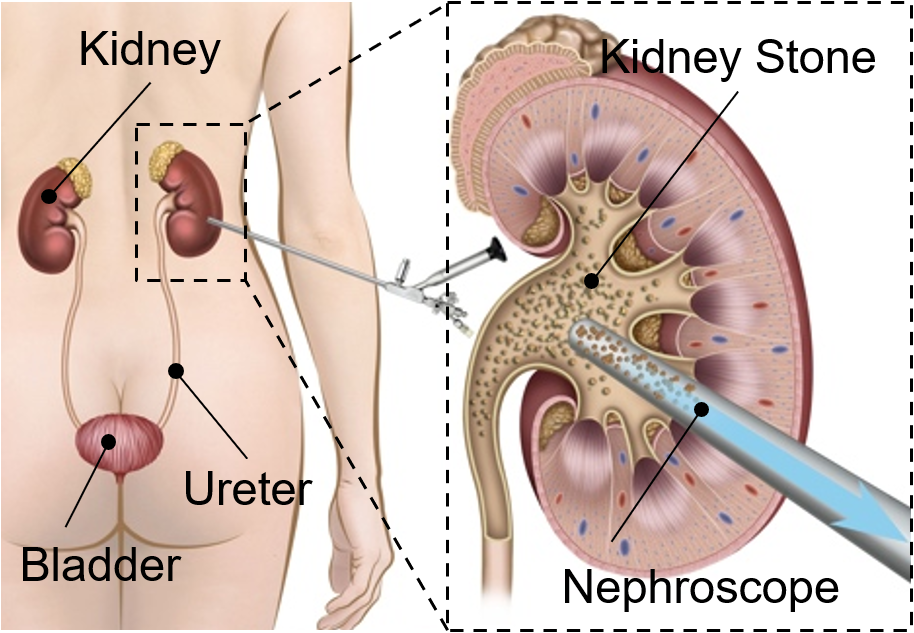}
    \caption{Illustration of Minimally Invasive Surgery procedures for removing large kidney stones with Percutaneous Nephrolithotomy.}
    \label{Fig_Background}
\end{figure}

\IEEEPARstart{P}{ercutaneous} nephrolithotomy (PCNL) is a well-established Minimally Invasive Surgery (MIS) procedure for accessing, locating, and removing larger kidney stones through a small incision on the back of the patient’s body\cite{de2015percutaneous}, as illustrated in Fig.\ref{Fig_Background}. 
However, certain challenges persist in PCNL operations; the needle incision point selection, the kidney stone visualization, as well as the visual guidance for the needle insertion task during surgical operations\cite{rodrigues2013kidney}. 

Ultrasound (US) imaging is a popular modality in PCNL surgery considering its low cost and radiation-free features\cite{jiang2021autonomous}. Paranawithana \textit{et al.} proposed an ultrasound-guided involuntary motion compensation of kidney stones; the surgeon checked the 2D ultrasound image of kidney stones and performed the puncture operation to locate the target calyx of the kidney \cite{paranawithana2018ultrasound}. Li \textit{et al.} designed a visualization interface to allow the surgeon to define the suitable percutaneous target by observing the ultrasound image \cite{li2018towards}. Tu \textit{et al.} proposed to use the real-time updated ultrasound image to reconstruct the soft tissue and the 3D model was updated for deformation correction \cite{tu2022ultrasound}. 
Except for the US image, Stoianovici \textit{et al.} developed a visualization interface and provided the surgeon with real-time x-ray fluoroscopy image for PCNL interventions\cite{stoianovici1998modular}. These studies indicate that echo image-guided procedures are technically possible. However, they demand a high level of hand-eye coordination from the surgeon to effectively manipulate the needle during the execution of the task.

Compared to conventional surgery, robot-assisted MIS techniques have been introduced into this surgical scenario considering their advantages of high positioning accuracy, improved dexterity, and reduced trauma and complications for faster post-recovery\cite{ferraguti2022augmented}. The use of robotic assistance for PCNL task execution has been extensively investigated during the past decades. For example, a portable modular surgical robot was originally designed and implemented for PCNL interventions\cite{stoianovici1998modular}. With the developed robotic system, the surgeon could remotely control the robotic manipulator with a joystick, allowing the surgeon to avoid radiation exposure. Wilz \textit{et al.} \cite{wilz2021constrained} developed a human-robot shared control framework for teleoperated percutaneous nephrolithotomy procedures training. The haptic feedback was integrated to guide the operator during needle insertion procedures. In addition, a robotic manipulator incorporating the ultrasound scanning system for PCNL procedures was implemented to assist the surgeon in completing the needle insertion task \cite{li2018towards}. Alongside the robotic assistance, the surgeon typically performs needle insertion, access to the target kidney stones, and removal operations based on visual feedback and guidance from the medical imaging, which still imposes a high workload on the surgeon both mentally and physically \cite{ferraguti2022augmented,wilz2021constrained}. 

Virtual Reality (VR) technique has been integrated into PCNL simulators to provide the user with immersive training experiences. For example, Sainsbury \textit{et al.} proposed a VR and haptic feedback-integrated surgical rehearsal system for PCNL task training, and improved surgical operation performance was observed\cite{sainsbury2022preoperative}. Similarly, Farcas \textit{et al.} developed a novel 3D immersive percutaneous renal access VR training platform, and comparison experiments were conducted to verify its superiority compared to the other two PCNL simulation platforms\cite{farcas2021simulation}. Furthermore, Augmented Reality (AR) enables users to visualize specific anatomical structures and obtain real-time intraoperative information regarding the patient's anatomy\cite{fu2023recent}. It combines various sources of information and presents them simultaneously, enhancing the surgical experience \cite{palumbo2022mixed}. For example, an AR interface was implemented to intuitively visualize the tumor position during the percutaneous puncture process in \cite{li2021towards}.
In a recent study by Wang \textit{et al.}, Optical See-Through Head-Mounted Display (OST-HMD) was investigated for providing surgeons with 3D visualization during the PCNL procedures for intraoperative guidance \cite{wang2022application}.
Experimental results indicated that the proposed framework achieved a targeting error of $3.1\pm2.9$ mm. Furthermore, when compared to routine B-ultrasound imaging feedback, the use of OST-HMD showed several advantages. It resulted in reduced time required to complete the puncture task, fewer puncture attempts were needed, and improved performance in terms of stone clearance rate.

The integration of an AR visualization interface and robotic assistance has demonstrated promising results in enhancing the surgeon's perception capabilities during the MIS procedures \cite{qian2019review}. 
For instance, Qian \textit{et al.} \cite{qian2019review} developed the \textit{“ARssist”} framework to help the first assistant to perform two specific tasks during robotic-assisted laparoscopic surgery. Utilizing the OST-HMD, showcased improved task completion efficiency, navigation consistency, and safety during instrument insertion tasks, particularly for inexperienced users. Additionally, it enhanced hand-eye coordination in tool manipulation tasks.
In \cite{lin2021arei}, an AR-based touchless teleoperation control interface was designed to intuitively control the distal position of a flexible robot for endoluminal intervention alternative to 2-D X-ray guidance. Moreover, in  \cite{tu2022ultrasound}, a Mixed Reality-based visualization interface for robotic cervical pedicle screw placement was developed, addressing the hand-eye coordination problem during the task execution. 

In this work, we address the challenges of performing robot-assisted PCNL tasks and propose a framework that integrates AR visualization and robotic assistance to improve the surgeon’s perception capability and task completion performance. The proposed framework makes the following contributions: 
\begin{itemize}
    \item It incorporates an intuitive AR visualization interface, utilizing the OST-HMD to display the patient's anatomical structures, assist with needle insertion path planning, and provide visual guidance during the insertion procedure; 
    \item It implements a robot-assisted navigation strategy that aids the surgeon in performing needle insertion, thereby enhancing the overall performance of the PCNL task.
\end{itemize}

The remainder of this paper is organized as follows. Section \ref{Section: Workflow} describes the workflow of the proposed framework. Section \ref{Section:Methodology} presents the details of the methodology. Section \ref{Section: System} gives the system details, the experimental protocol, and performance metrics. Section \ref{Section: Results} gives the experimental results and discussion. Finally, Section \ref{Section: Conclusion} concludes this work.

\section{Workflow of Proposed Framework}
\label{Section: Workflow}

\begin{figure*}
    \centering
    \includegraphics[scale=0.5]{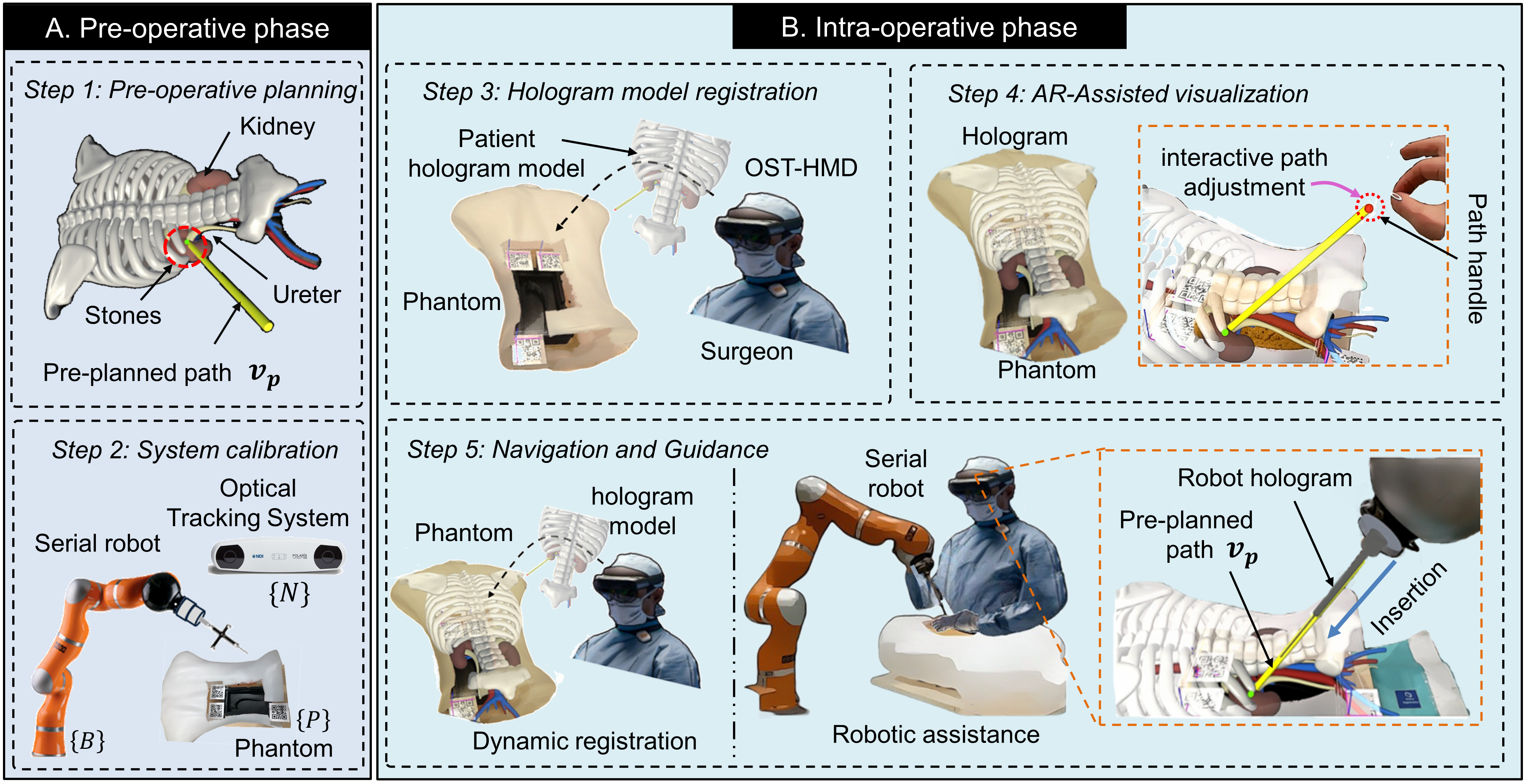}
    \caption{Workflow of the proposed framework for PCNL. In phase A, the CT scan is first conducted, segmentation and 3D reconstruction to retrieve the interested organs and the anatomy structure in \textit{“Step 1”}; A planned path for needle insertion with the kidney stone positions as the target is initialized (\textit{“Step 1”}); Afterwards, in \textit{“Step 2”} the system calibration and registration is implemented to find the coordinates transformation between system components. In phase B, the hologram model overlaps on the patient phantom after hologram to phantom registration (\textit{“Step 3”}); Interactive adjustment to obtain a collision-free needle insertion path (\textit{“Step 4”}); Human-Robot collaboration for needle insertion with AR visualization and robot guidance (\textit{“Step 5”}).}
    \label{Fig_framework}
\end{figure*}

\begin{figure*}
    \centering
    \includegraphics[scale=0.45]{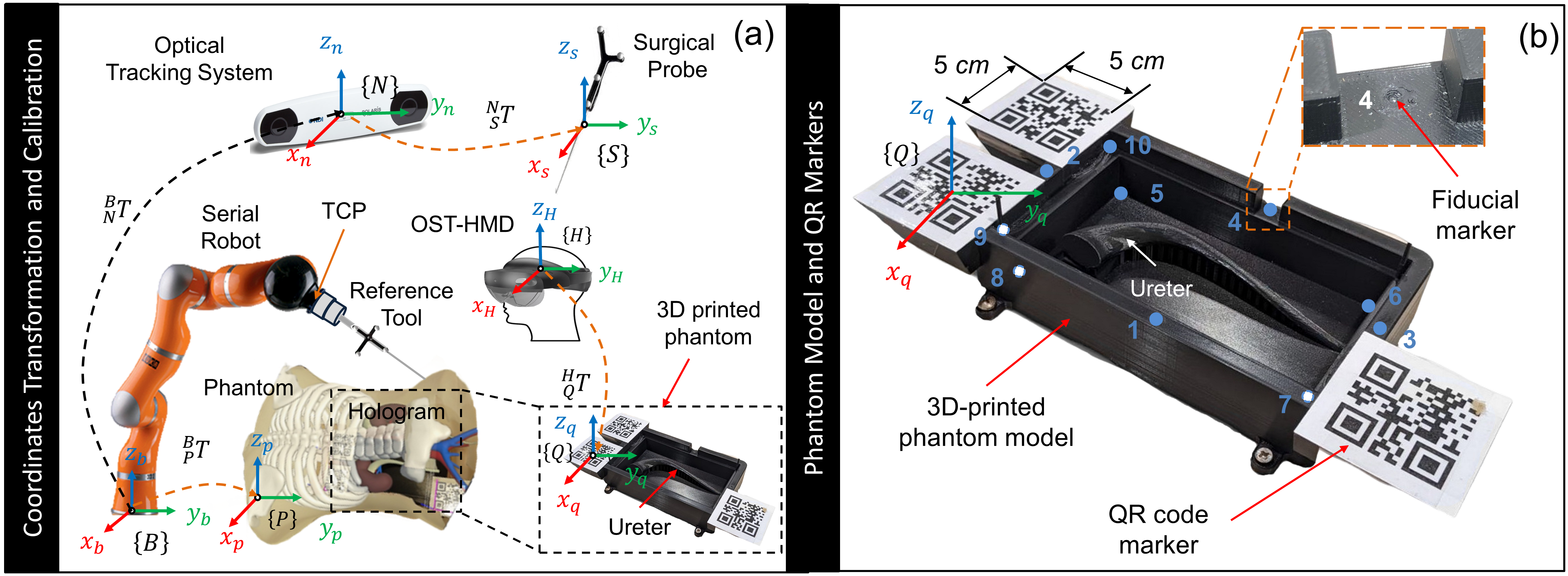}
    \caption{Coordinates transformation and system calibration. (a) System components coordinate transformation; (b) Calibration and registration with QR code and fiducial markers; The fiducial markers are on the 3D-printed phantom with a 1 mm diameter groove and evenly distributed over the phantom in spatial.}
    \label{Fig_Coordinates_System}
\end{figure*}

The workflow of the proposed framework encompasses both the pre-operative and intra-operative phases, which are further divided into multiple steps. Please refer to Fig.~\ref{Fig_framework} for a visual representation of these steps. The details of each phase and step are explained in the following.

\textit{(A) Pre-Operative Phase:} During this phase, the patient undergoes the Computer Tomography (CT) scan first. This scan captures detailed images of the targeted area's anatomy, allowing for a precise understanding of the patient's condition. Next, the patient’s anatomical structures are segmented from the scan data, allowing for the creation of a 3D model that represents the patient's anatomy, including the kidney, the stones, the ureter, the aorta, the surrounding structures, etc. Following this, the surgeon proceeds to define a pre-planned path, denoted as \textit{“Step 1”} and represented by $\bm{v_p}$, with the objective of reaching the target kidney stone position using the available 3D models. To calculate the coordinates transformation metrics between the system components, the calibration procedures between the serial robot, the optical tracking system, and the patient phantom model are implemented, as depicted in \textit{“Step 2”}.

\textit{ (B) Intra-Operative Phase:} In the proposed framework, an AR interface is adopted to provide surgeons with real-time visualization and guidance for task execution during the intra-operative phase. This is achieved by overlapping the hologram model onto the physical patient phantom model, namely, \textit{“Hologram model registration”}, as shown in \textit{“Step 3”}. Landmarks are typically adopted and attached to the patient's back during the CT scanning to obtain the corresponding relationship between the organs and landmarks, which is crucial for the following registration procedures.
Subsequently, in \textit{“Step 4”} the surgeon can visualize a holographic representation of the 3D model that is superimposed on the physical patient phantom. Moreover, the surgeon is allowed to interactively validate the intra-operative pre-planned path and obtain a collision-free path concerning the delicate anatomical structures by adjusting the pre-planned path orientation. Following that, the surgeon can perform the needle insertion task for PCNL with AR-assisted navigation and robotic assistance and guidance, as demonstrated in \textit{“Step 5”}.

\section{METHODOLOGY}
\label{Section:Methodology}

In the proposed framework, an external optical tracking system is employed for system component calibration, registration, and performance evaluation, see Fig.~\ref{Fig_Coordinates_System} (a). The coordinates transformation between the system components is illustrated. Specifically, ${ }_{N}^{B} \bm{{T}}$ is the coordinates transformation from the robot base $\{\bm{B}\}$ to the optical tracking system $\{\bm{N}\}$, ${ }_{P}^{B} \bm{{T}}$ is the transformation matrix from $\{\bm{B}\}$ to human phantom $\{\bm{P}\}$, ${ }_{S}^{N} \bm{{T}}$ represents the transformation matrix from $\{\bm{N}\}$ to the handheld surgical probe $\{\bm{S}\}$, and ${ }_{Q}^{H} \bm{{T}}$ is the one from the OST-HMD $\{\bm{H}\}$ to the QR code marker $\{\bm{Q}\}$, respectively.

\subsection{Systems Calibration and Registration}
In the proposed framework, the system calibration and registration should be implemented above all, which consists of three parts, namely, 1) \textit{Eye-to-Hand Calibration}; 2) \textit{Robot-to-Phantom Registration}; and 3) \textit{Hologram-to-Phantom Registration}. The details for each part are explained below.

\subsubsection{Eye-to-Hand Calibration}
First of all, the transformation matrix ${ }_{N}^{B} \bm{{T}}$ from the robot base $\{\bm{B}\}$ and the optical tracking system $\{\bm{N}\}$ is calculated with classical Eye-to-Hand calibration algorithm \cite{tsai1989new}. In detail, as depicted in Fig.~\ref{Fig_Coordinates_System} (a), a reference tool $\{\bm{R}\}$, attached to the robot and tracked by the optical tracking system, is adopted for calibration. According to the geometric relationship, we have:
\begin{equation}
   { }_{N}^{B} \bm{T}={ }_{T}^{B} \bm{T} *{ }_{R}^{T} \bm{T} *{ }_{N}^{R} \bm{T}
   \label{calibration_1}
\end{equation}
where ${ }_{R}^{T} \bm{T}$ denotes the transformation matrix from the Tool Centre Point (TCP) point of the robot to the reference tool $\{\bm{R}\}$, ${ }_{N}^{R} \bm{T}$ is the transformation matrix from the reference tool $\{\bm{R}\}$ to the optical tracking system $\{\bm{N}\}$. With two different robot configurations within its workspace, we have:
\begin{equation}
\left\{\begin{array}{c}
{ }_{R}^{T} \bm{T}={ }_{T}^{B} \bm{T}_{1}^{-1} *{ }_{N}^{B} \bm{T} *{ }_{N}^{R} \bm{T}_{1}^{-1} \\
\\
{ }_{R}^{T} \bm{T}={ }_{T}^{B} \bm{T}_{2}^{-1} *{ }_{N}^{B} \bm{T} *{ }_{N}^{R} \bm{T}_{2}^{-1}
\end{array}\right.
\label{EqualMatrix}
\end{equation}
Eq. \eqref{EqualMatrix} can be further simplified as: $\bm{AX}=\bm{XB}$, where $\bm{A}={ }_{T}^{B} \bm{T}_{2} *{ }_{T}^{B} \bm{T}_{1}^{-1}$, $\bm{{B}}={ }_{N}^{R} \bm{T}_{2}^{-1} *{ }_{N}^{R} \bm{T}_{1}$, and $\bm{X}={ }_{N}^{B} \bm{T}$. Afterwards, changing the robot to different configurations within its workspace for ${n}$ times, several groups of mathematical equations in the format of Eq. \eqref{EqualMatrix} can be obtained and Tsai's algorithm \cite{tsai1989new} is then adopted for solving the transformation matrix, ${ }_{N}^{B} \bm{T}$.

\subsubsection{Robot-to-Phantom Registration}
To provide the human operator with intuitive visualization of the pre-planned path and robotic assistance during PCNL procedures, the transformation matrix ${ }_{P}^{B} \bm{T}$ between the robot base frame $\{\bm{B}\}$ and the phantom frame $\{\bm{P}\}$ needs to be estimated. In Fig.\ref{Fig_Coordinates_System} (a), 10 spatial distributed fiducial markers are collected on the 3D printed phantom model, which is denoted by $\bm{M^P}=\left\{M_i^P \mid i=1, \ldots 10\right\}$ and these positions are known with respect to  (w.r.t) the phantom model coordinate system $\{\bm{P}\}$. Since the transformation matrix ${ }_{N}^{B} \bm{T}$ is obtained with Eq. \eqref{EqualMatrix}, the handheld surgical probe, which is tracked by the tracking system, is used to acquire the position of the fiducial markers w.r.t the robot base $\{\bm{B}\}$, recorded as $\bm{M^B}=\left\{M_i^B \mid i=1, \ldots 10\right\}$. To improve accuracy, the surgical probe tip position is calculated by an average of 50 times data acquisition at each fiducial marker position. Afterwards, the Singular Value Decomposition (SVD) algorithm \cite{arun1987least} is adopted to calculate ${ }_{P}^{B} \bm{T}$.

\subsubsection{Hologram-to-Phantom Registration} 
To intuitively provide the surgeon with patient anatomical structure and organ model, the transformation matrix ${ }_{Q}^{H} \bm{T}$ for registration between the OST-HMD $\{\bm{H}\}$ and the phantom hologram model $\{\bm{P}\}$ needs to be estimated. This involved a one-time setup where the user, wearing the HoloLens 2 headset, simply needed to detect these three QR code markers by looking at them one by one. The whole process for this procedure takes approximately 20 seconds. To achieve accurate registration results, three QR markers, with 5x5 cm in size for each marker, are attached to the phantom model and they can be detected by the front camera of the OST-HMD at a distance of roughly 20 cm \footnote{\url{https://github.com/GlitchEnzo/NuGetForUnity/releases}}. Typically, the larger and closer the HoloLens is to the QR Code, the faster it can be recognized. And it should be bigger than 2x2 cm so that the front camera of HoloLens 2 can detect it quickly and robustly. In addition, good lighting conditions help detect QR markers.

As illustrated in Fig.\ref{Fig_Coordinates_System} (a) and (b), the positions of these three markers w.r.t the phantom coordinate system are known, which are represented as $\bm{Q^{ph}}=\left\{Q_i^{ph} \mid i=1,2,3\right\}$. Considering the different perspectives of the OST-HMD camera concerning the user's eye, the \textit{Hologram-to-Phantom Registration} step is conducted before the experiment for each user. After these QR code markers' positions $\bm{Q^{ho}}=\left\{Q_i^{ho} \mid i=1,2,3\right\}$ w.r.t the OST-HMD coordinate system $\{\bm{H}\}$ are acquired once they are recognized by the front camera of the OST-HMD and the SVD is adopted to compute the matrix ${ }_{Q}^{H} \bm{T}$ afterwards. 

Once the Hologram-to-Phantom registration was done, the hologram model aligned with the physical phantom model, and the markers were not no longer necessary for visibility. After the registration was implemented, the coordinate transformation was established, enabling the human operator to view the holographic model from different perspectives while the hologram model remained stationary.

\subsection{Augmented Reality-Assisted Visualization and Navigation}

In the left panel of Fig.\ref{Fig_System}, the AR visualization interface is implemented to intuitively display the patient's anatomical structure and organ holographic model, including kidneys, aorta, ureters, etc., during the surgical operations. A pre-planned needle insertion reference path $\bm{v_p}$ in the proposed framework is visualized from OST-HMD, which is overlapped with the physical phantom model. Moreover, considering the inherent risks of potential collisions between the needle and surrounding anatomical structures during the task execution of PCNL procedures, a collision-free insertion path should be pre-planned. Consequently, an interactive path adjustment interface, has been implemented to enable the operator to redefine the needle insertion path. Specifically, as illustrated in \textit{“Step 4”} of Fig.\ref{Fig_framework}, a path handle is attached to the endpoint of the pre-planned path. The human operator can easily adjust the initially established path around the target point with respect to the coordinate frame of the kidney stone by manipulating the "path handle" endpoint. Once the necessary adjustments to the pre-planned path have been made to obtain a collision-free insertion path, the human operator can disable the \textit{interactive path adjustment} by pressing the button on the GUI interface.

Furthermore, to transparently guide the human operator during both alignment and needle insertion phases, the holographic needle model is also visualized, and its pose is updated in real-time derived from the forward kinematics formulation of the serial robotic manipulator. More details related to the hologram model of the serial robot and the holographic robot configuration update can be found in our previous work\cite{fu2023augmented}.

\subsection{Human-Robot Collaboration Strategy for PCNL}
In the right panel of Fig.~\ref{Fig_System}, the proposed shared control strategy for human-robot collaboration is composed of three manipulation steps. The details are explained as follows:

\begin{figure*}
    \centering
    \includegraphics[scale=0.55]{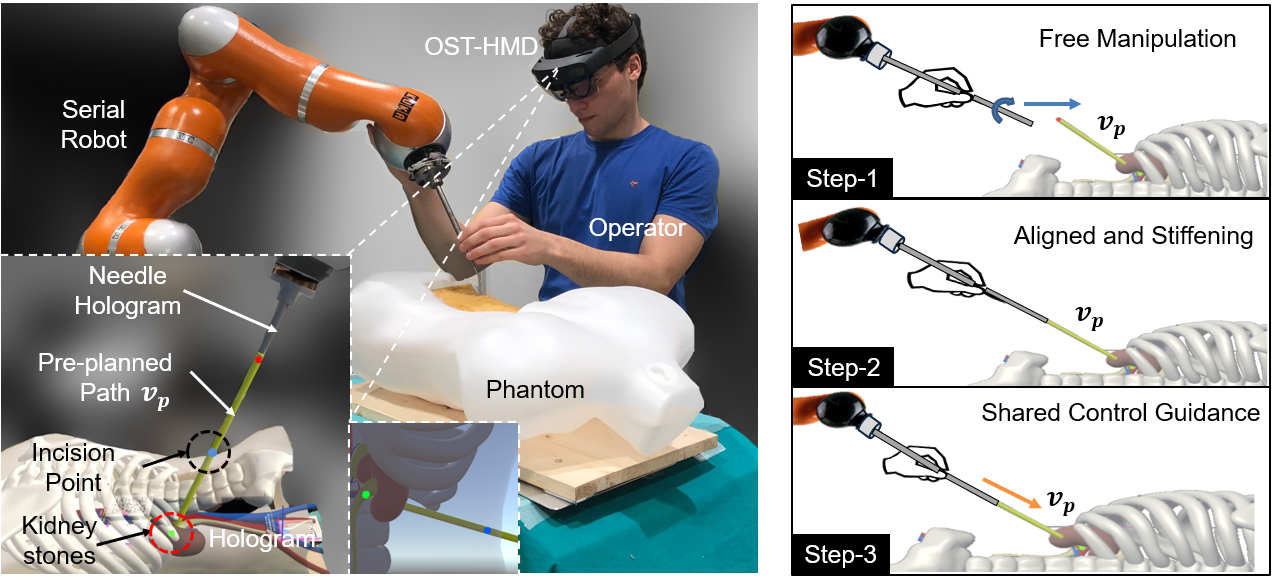}
    \caption{System implementation and proposed control strategy. Left: System components of the proposed AR and human-robot collaboration control framework for PCNL; Right: Human-robot collaboration strategy for Robot-Assisted PCNL task, including 3 sub-steps.}
    \label{Fig_System}
\end{figure*}

\textbf{\textit{Step-1:}} (Free Manipulation for Alignment) During this stage, the robot is working with the Gravity Compensation (GC) modality \cite{fu2023augmented} that allows the surgeon to manipulate freely and to align the needle to the pre-defined path, which is intuitively visualized by the OST-HMD.

\textbf{\textit{Step-2:}} (Aligned and Stiffening of the Robot) When the alignment between the pre-planned path and the real needle is implemented, the robot can be stiffened by pressing the “Aligned” button on the GUI from the OST-HMD interface. Afterwards, the robot could only move along the pre-planned path for insertion with respect to the robot end effector coordinate system.

\textbf{\textit{Step-3:}} (Shared Control Guidance for Puncture) Once the robot is stiffened, the robot can assist the surgeon in performing needle puncture only in the pre-planned direction. This could relieve both the physical and mental workload from the surgeon, thus, allowing the surgeon only focus on the puncturing and kidney stone removal task.
In \textit{Step-3}, the Cartesian impedance controller is implemented for assistance during PCNL procedures while simultaneously ensuring safety and robotic system compliance. The contact force $\bm{F_{{ext }}}$ between the robot and human operator is calculated by:
\begin{equation}
\bm{F_{ext}}=\bm{{M\ddot{\tilde{x}}}}+\bm{K_{c} \tilde{x}}+\bm{D_{c} \dot{\tilde{x}}}
\label{Impedance Control}
\end{equation}
where $\bm{\tilde{x}}= \bm{x_{d e s}}-\bm{x_{m s r}} $, $\bm{x_{des}}$ and $\bm{x_{msr}} \in \mathbb{R}^m$ are the desired and measured current robot position in Cartesian coordinates of the robotic system. $\bm{\dot{\tilde{x}}}$ and $\bm{\ddot{\tilde{x}}} \in \mathbb{R}^m$ are the measured velocity and acceleration. Considering the low-speed motion and acceleration noise in the real scenario, the acceleration term $\bm{{M\ddot{\tilde{x}}}}$ is usually neglected for simplification of the Eq.\eqref{calibration_1}. $\bm{K_{c}}$ and $\bm{D_{c}} \in \mathbb{R}^{m \times m}$ are the positive definite stiffness and damping parameters. $\bm{D_{c}}$ can be calculated by critical damping law
$2 * \xi \sqrt{{K_{c}}}$, where $\xi \in[0,1]$ is the damping coefficient and typically the critical damping value of $\xi = 0.707$ is adopted. Furthermore, the dynamics model of a $n$-Degree-of-Freedoms (DoFs) robotic manipulator in the Cartesian space is formulated as:
\begin{equation}
\bm{M(x)} \bm{\ddot{x}}+\bm{C(x, \dot{x}) \dot{x}}+\bm{F_g(x)}-\bm{F_{ext}}=\bm{F_{cmd}}
\label{Cartesian_Dynamic}
\end{equation}
where $\bm{M}$, $\bm{C} \in \mathbb{R}^{m \times m}$ are the symmetric and positive definite mass, Coriolis and Centrifugal matrix, respectively. $\bm{F_{{g}}}$, $\bm{F_{{ext }}}$, and $\bm{F_{{cmd}}} \in \mathbb{R}^m$ are the gravity, external, and commanded wrench in Cartesian space. The joint torque to control the robot is calculated by $\bm{\tau_{{ext }}}=\bm{J(q)}^T \bm{F_{ext}}$, $\bm{G(q)}=\bm{J(q)}^T \bm{F_g(x)}$, $\bm{\tau_{c m d}}=\bm{J(q)}^T \bm{F_{c m d}}$, respectively. $\bm{\tau_{ext}}$, $\bm{\tau_{ext}}$, and $\bm{G(q)} \in \mathbb{R}^n$ are the vectors of external force, robot command force, and gravity force torques, respectively. $\bm{J(q)} \in \mathbb{R}^{m \times n}$ is the Jacobian matrix and $\bm{q} \in \mathbb{R}^{n}$ is the joint vector.

\section{Systems and Experiments}
\label{Section: System}

\subsection{System Implementation}
The proposed framework's hardware system and software communication protocol are given in Fig.~\ref{Fig_Communication}, mainly including:

\begin{itemize}
\item An Optical See-Through Head-Mounted Display (OST-HMD) HoloLens 2 (Microsoft, WA, USA) is adopted for holographic model visualization, incorporates Universal Windows Platform (UWP) from Unity3D and Mixed Reality Toolkit (MRTK); 

\item A 7-DoFs serial robotic manipulator (LWR 4+, KUKA, Germany) is adopted for the needle insertion process; 

\item An external optical tracking system, Polaris Vicra, (NDI, Northern Digital Inc., Canada);

\item To facilitate data exchange, the Ubuntu 16.04 desktop incorporates the Robot Operating System (ROS) kinetic version is exploited and the robot is controlled with the FastResearchInterface (FRI) Library from KUKA company. As shown in Fig.~\ref{Fig_Communication}, the PC-1 connects to the serial robot using the User Datagram Protocol (UDP) and controls the robot with the FRI Library. On the other hand, the PC-2, equipped with Windows 10, i7-9750H CPU running at 2.60GHz with 16GB RAM, establishes connections with both the optical tracking system via USB and the ROS network through the Ethernet cable.

\begin{figure}
    \centering
    \includegraphics[scale=0.42]{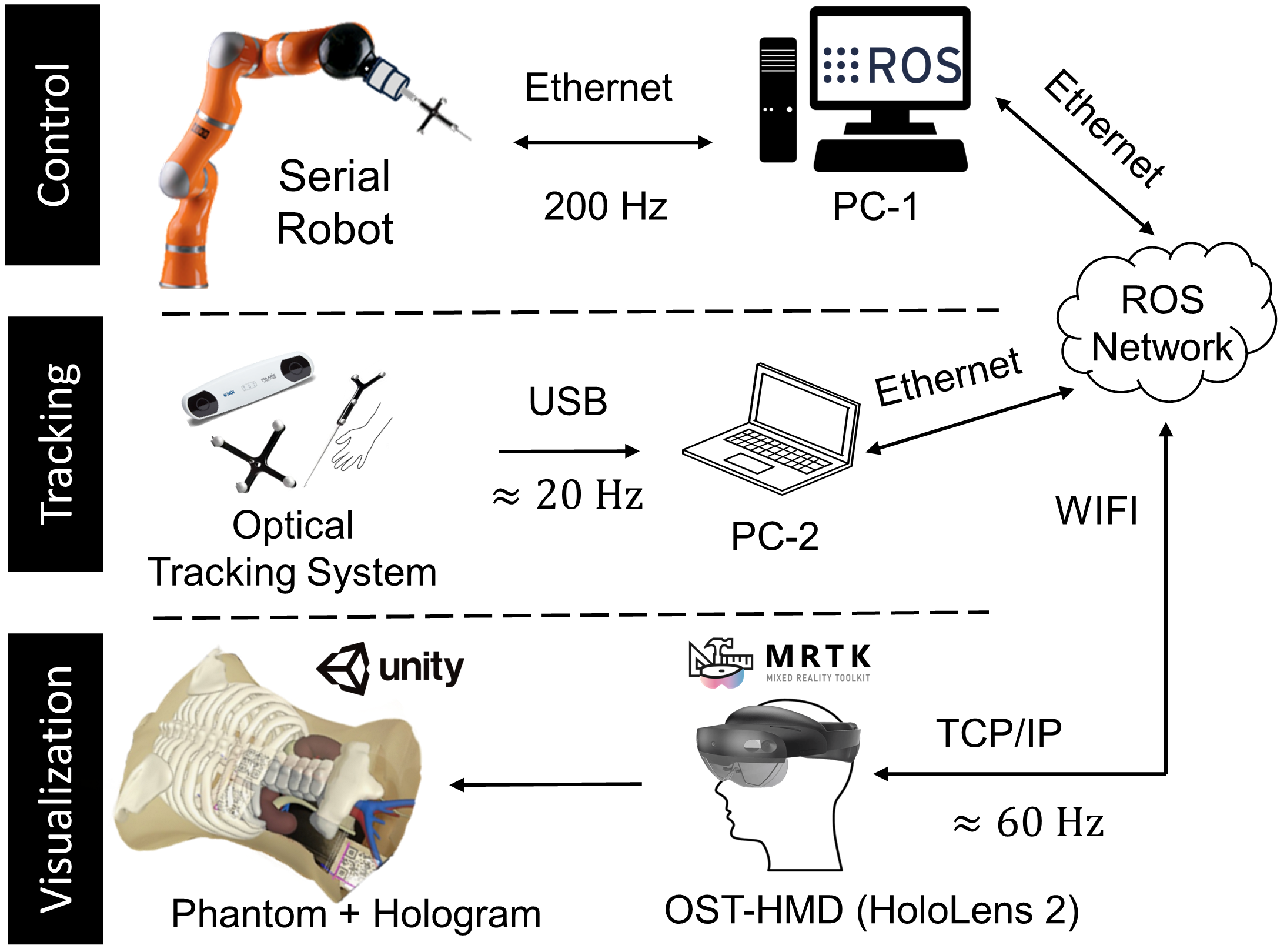}
    \caption{Hardware components and software communication protocol of the proposed system, including \textit{“Control”},\textit{“Tracking”}, and \textit{“Visualization”} layers. TCP/IP represents Transmission Control Protocol/Internet Protocol.}
    \label{Fig_Communication}
\end{figure}

\end{itemize}

\subsection{Experimental Protocol}

\subsubsection{Experimental Setup} 
\begin{figure}
    \centering
    \includegraphics[scale=0.42]{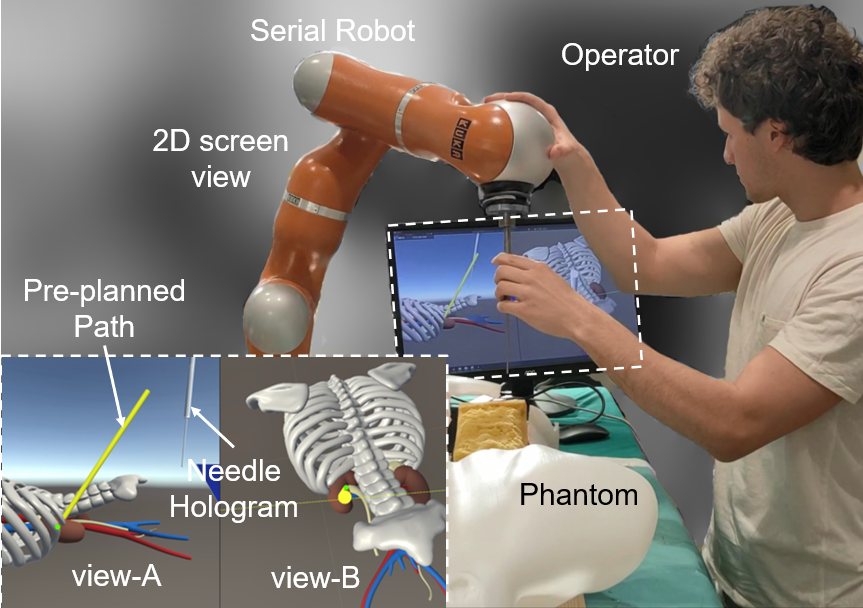}
    \caption{Experimental setup with 2D screen visualization interface for the PCNL task. The user manipulates the robot and aligns the needle hologram to the pre-planned path by observing both view-A and view-B on the screen.}
    \label{Fig_2D}
\end{figure}

To verify the performance of the proposed framework during the task execution, four different modalities are implemented. The details for each experimental setup are described as follows:
\begin{itemize}
    \item \textbf{SM:} 2D Screen-based visualization + Manual Control; The participants manually manipulated the needle that was mounted on the robot end effector to align to the pre-defined path for a puncture, which was visualized by a 2D monitor, as shown in Fig.\ref{Fig_2D}. Afterwards, the participants did the needle insertion task to the target position (see \textit{“Step 5”} in Fig.\ref{Fig_System}) manually without robot assistance. During the task execution process, the user can adjust the view of the virtual environment.
    \item \textbf{SG:} 2D Screen-based visualization + Robot Guidance; The participants were required to align the needle to the pre-planned path by observing the 2D screen. And, robotic assistance was provided in the following needle insertion procedure.
    \item \textbf{AM:} AR-based visualization + Manual Control; In this setup, participants wore the OST-HMD to observe and align the needle tip to the pre-planned path and did manual insertion after alignment for the PCNL task without robot assistance.

    \item \textbf{AG:} AR-based visualization + Robot Guidance; The participants observed the pre-planned path guidance from the OST-HMD and performed the needle insertion with the robot guidance using the proposed HRC strategy.
\end{itemize}

Specifically, for the setups of \textbf{SM} and \textbf{AM}, the robot was running with the Gravity Compensation (GC) model during the whole procedure allowing the user to freely manipulate the needle. For \textbf{SG} and \textbf{AG}, the robot stiffness $\bm{K_{c}}$ along the needle insertion direction and the other two directions are set as 0 N/m and high stiffness with $4 \times 10^3$ N/m, respectively, and the rotational stiffness is 200 Nm/rad. Moreover, the damping parameter is the critical value of stiffness. Thus, after alignment, the robot can only move along the pre-planned path for needle insertion. The experimental protocol was approved by the ethics committee from Politecnico di Milano, Italy (No.2023-5069).

\subsubsection{User Evaluation} 
$P = 14$ healthy participants were invited to participate in the experiments (8 male and 6 female, aged between 22-27 with Avg = 24.2, Std = 1.21). All of them gave their informed consent about the steps and contents of the experiments. 
Before the experiments, the system developer, who was the biomedical engineering researcher, provided comprehensive instructions to all the participants to be familiar with the OST-HMD device, the robot manipulation, and the details of the experiment procedures. Following that, each participant was allocated approximately 30 minutes for the hands-on practice. Moreover, all the participants practiced the needle insertion task using the same trajectory as the one in the following experiments, across all four experimental setups.

The order of these 4 different experimental setups was randomly conducted with different participants to avoid unfair comparison. All of them performed the PCNL task with 4 different modalities, each modality with 3 repetitions. After finishing the task with each setup, the participants were asked to fill in a NASA-TLX\cite{hart2006nasa} questionnaire, including \textit{“Mental Demand”}(MD),\textit{“Physical Demand”}(PD), \textit{“Temporal Demand”}(TD), \textit{“Performance”}(PE), \textit{“Effort”}(EF), and \textit{“Frustration”}(FR) items. It should be noted that the individual raw scores for these six items are rated on a scale ranging from 0 to 20. The overall score is derived by summing up all these items and is further scaled from 0 to 100 for better clarity and interpretation.

\subsubsection{Performance Metrics}
The following metrics are adopted to verify the performance of the proposed framework, consisting of system calibration and task completion performances. 

(i) Performance metrics related to calibration and registration, including \textit{Eye-to-Hand} calibration, \textit{Robot-to-Phantom Registration}, and \textit{Hologram-to-Phantom Registration}.
\begin{itemize}
\item To quantitatively evaluate the \textit{Eye-to-Hand calibration} error between the robot $\{\bm{B}\}$ and optical tracking system $\{\bm{N}\}$, the Root Mean Squared Error (RMSE) on the 3d-printed phantom model fiducial marker positions in Fig.\ref{Fig_Coordinates_System}(b), is calculated as: 
\begin{equation}
RMSE_{E2H} = \sqrt{\frac{1}{M} \sum_{m=1}^M E_m^2}
\end{equation}
where $E_m=\left\|\bm{P}_m^N-\bm{P}_m^B\right\|$ represents the difference of the $m$-th with $m=1,2, \ldots, M$ positions measured by the optical tracking system $\bm{P_}m^N$ and acquired from a physical serial robot $\bm{P_}m^B$, respectively.

\item The accuracy of \textit{Robot-to-Phantom Registration} is evaluated by calculating the RMSE of the fiducial marker’s positions obtained from the phantom coordinate system $\{\bm{P}\}$ and the robot $\{\bm{B}\}$, respectively, and calculated by:
\begin{equation}
RMSE_{R2P} = \sqrt{\frac{1}{N} \sum_{n=1}^N E_n^2}
\end{equation}
where $E_n=\left\|\bm{P}_n^P-\bm{P}_n^B *{ }_P^B \bm{T}\right\|$ denotes the error on the $n$-th with $n=1,2, \ldots, N$ fiducial marker positions (see Fig.\ref{Fig_Coordinates_System}(b)), which is calculated by the difference between position on the phantom $\bm{P}_n^P$ and the position calculated by \textit{Robot-to-Phantom} transformation $\bm{P}_n^B *{ }_P^B \bm{T}$.

\item The accuracy of \textit{Hologram-to-Phantom Registration} is calculated by the $RMSE_{H2P}$ value between the hologram and the physical fiducial marker positions, represented as 
\begin{equation}
RMSE_{H2P} = \sqrt{\frac{1}{J} \sum_{j=1}^J E_j^2}
\end{equation}
where $E_j$ denotes the distance of the $j$-th, $j=1,2,\ldots, J$, holographic and physical position on the ureter phantom mode, which is calculated by:
\begin{equation}
 E_j=\left\|_N^P \bm{T} * \bm{P}_j^{{ver-p}}-{ }_N^H \bm{T} * \bm{P}_j^{{ver-h}}\right\|   
\end{equation}
where $\bm{P}_j^{{ver-p}}$ and $\bm{P}_j^{{ver-h}}$ are the $j$-th vertice position on the physical and holographic ureter phantom model. ${_N^P}\bm{T}$ and ${_N^H}\bm{T}$ are the transformation matrix from the phantom $\{\bm{P}\}$ and OST-HMD $\{\bm{H}\}$ to the optical tracking system $\{\bm{N}\}$, respectively. A detailed description of how the hologram-to-phantom registration accuracy is evaluated can be found in our previous work\cite{palumbo2022mixed}. 

\end{itemize}

\begin{figure}
    \centering
    \includegraphics[scale=0.4]{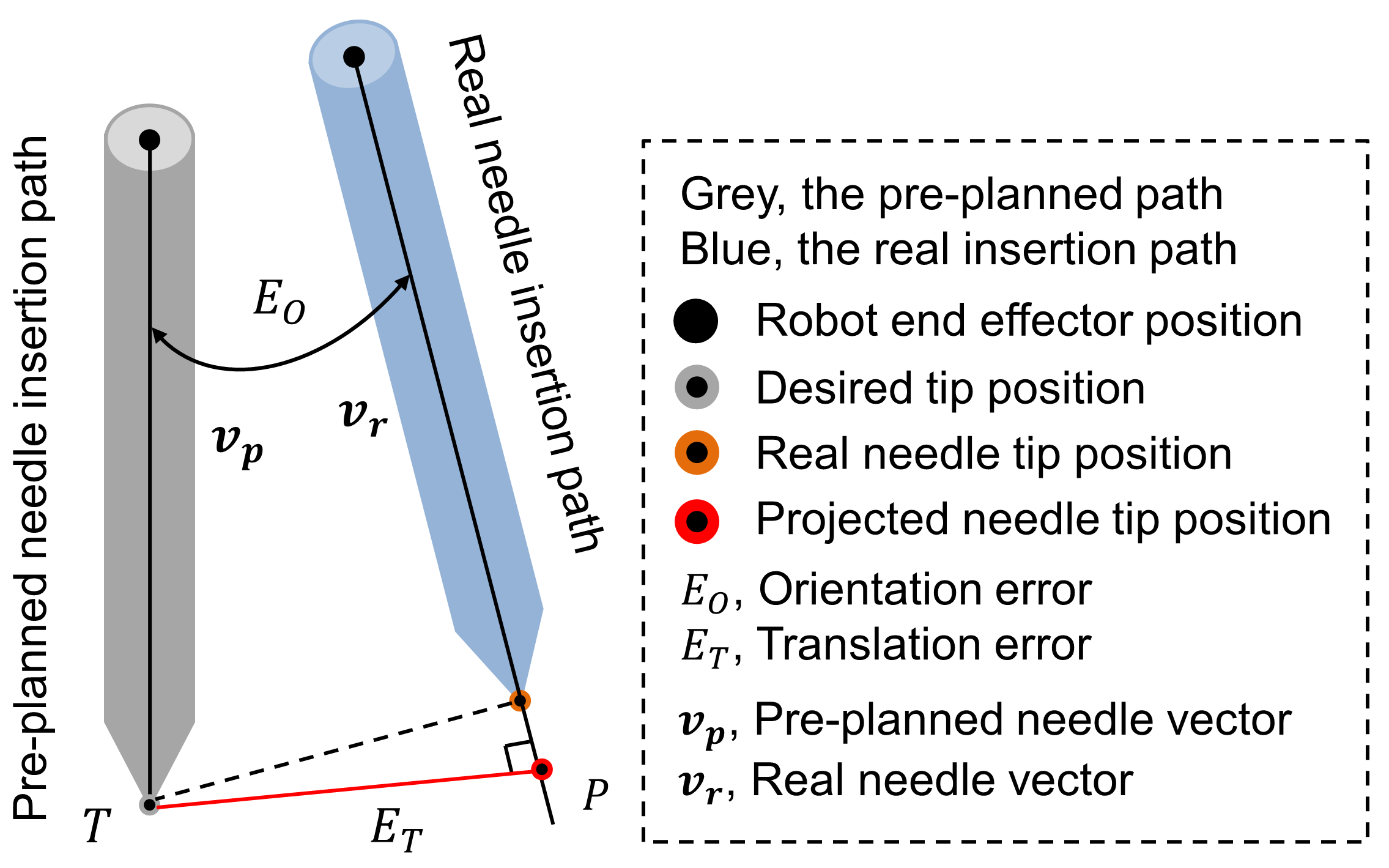}
    \caption{Translation error $E_T$ and orientation error $E_O$ calculation between the pre-planned and real needle insertion paths\cite{benmahdjoub2023evaluation}.}
    \label{Fig_Error_Cal}
\end{figure}

(ii) Performance metrics related to task completion, including the translation error, $E_T$[mm] and orientation error, $E_O$ [$\degree$], time cost, $T_{tot}$[s], and NASA-TLX user workload evaluation.

\begin{itemize}
\item The translation error $E_T$[mm] and orientation error $E_O$ [$\degree$] for the PCNL task between the reference path and the actual one, as shown in Fig.\ref{Fig_Error_Cal}, which are calculated by: 

\begin{equation}
\begin{aligned}
& E_O=\cos ^{-1}\left(\frac{\boldsymbol{v}_{\boldsymbol{p}} \cdot \boldsymbol{v}_{\boldsymbol{r}}}{\left\|\boldsymbol{v}_{\boldsymbol{p}}\right\| \cdot\left\|\boldsymbol{v}_{\boldsymbol{r}}\right\|}\right) \\
& P=\left\|\boldsymbol{v}_{\boldsymbol{p}}\right\| \cdot \cos \left(E_O\right) \cdot \frac{\boldsymbol{v}_{\boldsymbol{r}}}{\left\|\boldsymbol{v}_{\boldsymbol{r}}\right\|} \\
& E_T=\|\bm{P}-\bm{T}\|
\end{aligned}
\end{equation}
where $\bm{v_p}$ and $\bm{v_r}$ are the pre-planned and real needle vectors, from the robot end effector to the needle tip. $\bm{T}$ and $\bm{P}$ are the desired and projected needle tip positions on the real needle tip, in Fig.\ref{Fig_Error_Cal}. 

\begin{figure*}
    \centering
    \includegraphics[scale=0.54]{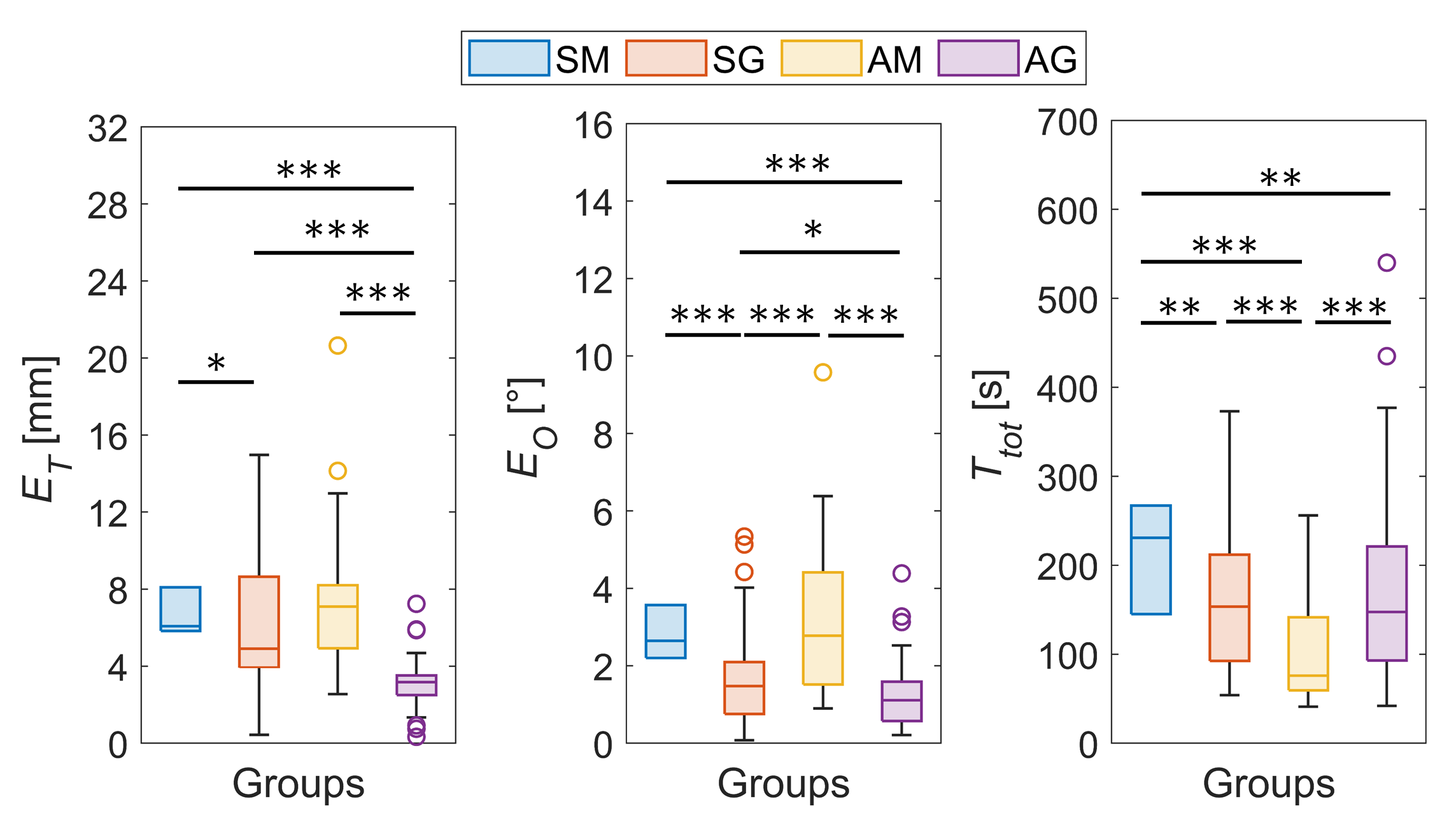}
    \caption{Task completion performance metrics from all four modalities, including translation errors $E_T$, orientation errors $E_O$, and time cost $T_{tot}$. ($ *, p<0.05 ; * *, p<0.01 ; * * *, p<0.001 $)}
    \label{Fig_Error}
\end{figure*}

\item The total time cost, $T_{tot}$[s] for completing the PCNL task, including the needle alignment and insertion tasks; 
\end{itemize}

(iii) To access the perceived cognitive workload of the human operator during the tasks, the NASA-TLX questionnaire evaluation was employed across all modalities, where a lower score indicates lower workload task completion. In addition, the significant differences in both the task performance errors and cognitive workload results are calculated to access the qualitative performance of the experimental results across different groups. The non-parametric statistical significance test, the Wilcoxon Rank-Sum test is adopted for statistical analysis of the comparison results. A significant difference can be assessed with $p$-value $<0.05$.

\section{Experimental Results and Discussion}
\label{Section: Results}

\begin{table*}
\caption{Task Completion Comparison Results with Median and SD values}
\centering
\begin{tabular}{|c|l|l|l|l|}
\hline
\diagbox{Metrics}{Groups} & \multicolumn{1}{c|}{SM}   & \multicolumn{1}{c|}{SG}  & \multicolumn{1}{c|}{AM}  & \multicolumn{1}{c|}{AG} \\ 
\hline
\multicolumn{1}{|c|}{$E_T$[mm]}  & \multicolumn{1}{l|}{$6.08 \pm 1.00$} & \multicolumn{1}{l|}{$4.90\pm 3.15$}  & \multicolumn{1}{l|}{$7.10 \pm 3.57$}  & \multicolumn{1}{l|}{$ \bm{3.17 \pm 1.36}$} \\ \hline

\multicolumn{1}{|c|}{$E_O$[\degree]}  & \multicolumn{1}{l|}{$2.65 \pm 0.59$}  & \multicolumn{1}{l|}{$1.47 \pm 1.36$}  & \multicolumn{1}{l|}{$2.78 \pm 1.80$}  & \multicolumn{1}{l|}{$ \bm{1.09 \pm 0.88}$} \\ \hline

\multicolumn{1}{|c|}{$T_{tot}$[s]}  & \multicolumn{1}{l|}{$231.0 \pm 50.0$}  & \multicolumn{1}{l|}{$153.5 \pm 82.0$}  & \multicolumn{1}{l|}{$\bm{76.0 \pm 55.0}$}  & \multicolumn{1}{l|}{$147.5 \pm 109$} \\ \hline
\end{tabular}
\label{table:error}
\end{table*}

\begin{figure*}
    \centering
    \includegraphics[width=\textwidth]
    {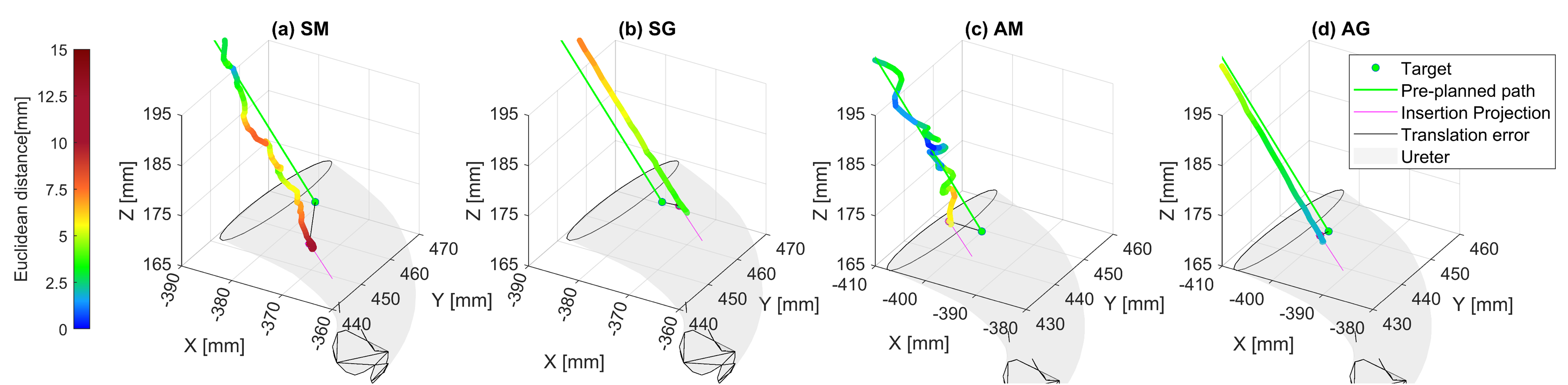}
    \caption{The translation error distance between the needle tip and the pre-planned path during the insertion phase with four different experimental group setups. From left to right panels: (a) Screen-based manual insertion (SM); (b) Screen-based robotic guidance (SG); (c) AR-based manual insertion (AM); (d) AR-based robotic guidance (AG). The targeting errors with these four setups are $9.87$ mm, $2.70$ mm, $5.83$ mm, and $1.61$ mm, respectively.}
    \label{Fig_Insetion_Error}
\end{figure*}

\subsection{System Calibration and Registration}
First of all, the \textit{Eye-to-Hand calibration} between the optical tracking device $\{\bm{N}\}$ and the robot base $\{\bm{B}\}$ was implemented by changing the robot to 80 different configurations within its workspace. To achieve higher calibration accuracy, the posture of the robot within its workspace between two adjacent samples should vary significantly. The \textit{Eye-to-Hand calibration} results were evaluated firstly, which provided an RMSE value of $RMSE_{E2H}$ with $1.44$ mm at 4 fiducial markers positions on the 3D printed phantom model in Fig.\ref{Fig_Coordinates_System}. 

After the \textit{Eye-to-Hand calibration}, the transformation matrix between the robot base $\{\bm{B}\}$ and the phantom model $\{\bm{P}\}$ is calculated using the SVD algorithm by collecting the 4 fiducial marker positions on the phantom w.r.t $\{\bm{B}\}$ and $\{\bm{P}\}$, respectively (see Fig.\ref{Fig_Coordinates_System}). To improve the accuracy of the collected fiducial marker’s position, each fiducial marker position is calculated as the average value of 50 consecutive frames of data from the NDI tracking device (the frequency of data acquisition is approximately 20 Hz). The robot-to-phantom registration error $RMSE_{R2P}$ is reported as $1.49$ mm and the median error and SD is $1.37\pm0.65$ mm.

Afterwards, the \textit{Hologram-to-Phantom Registration} result is calculated with the 20 spatial distributed vertices positions on the ureter model, as shown in Fig.\ref{Fig_Coordinates_System} (b). The real vertices positions w.r.t the phantom $\{\bm{P}\}$ are known and the holographic vertices positions are calculated by multiplying the transformation matrix ${ }_{Q}^{H} {\bm{T}}$. The registration result gives an RMSE error $RMSE_{H2P}$ of $2.44$ mm, with Median and SD values of $2.46\pm0.14$ mm.

\subsection{Task Completion Performance Comparison Results}

The participants’ task completion performance with all four groups of experimental setups is depicted in Fig.\ref{Fig_Error}, including the translation errors $E_T$, the orientation errors $E_O$, and the total time cost $T_{tot}$. For further clarity, the comparison of the experimental results for all performance metrics is listed in Table \ref{table:error}. In detail, the left panel of Fig.\ref{Fig_Error} gives the results of the translation error $E_T$ and shows that the proposed AG strategy achieves the minimum errors in the PCNL task, with a Median and SD value of $3.17\pm1.36$ mm. Furthermore, a significant difference between the SM and SG groups is observed with ${p}<0.05$ ($6.08\pm1.00$ mm vs $4.90\pm3.15$ mm), as well as between the AM and AG with ${p}<0.001$ ($7.10\pm3.57$ mm vs $3.17\pm1.36$ mm). Hence, the translation error is reduced with the robotic guidance in both screen-based (SM vs SG), and AR-based (AM vs AG) visualization modalities compared with manual insertion. In addition, the significant difference between the SG and AG ($4.90\pm3.15$ mm vs $3.17\pm1.36$ mm) illustrates the improved performance with AR guidance compared to the screed-based visualization, with ${p}<0.001$.

The middle panel of Fig.\ref{Fig_Error} demonstrates the orientation error $E_O$, in which the AG setup gives the minimum errors across all the setups, with a Median and SD value of $1.09\pm0.88\degree$, as depicted in Table \ref{table:error}. Besides, a significant difference between the SM and SG ($2.65\pm0.59\degree$ vs $1.47\pm1.36\degree$) is observed with ${p}<0.001$. Also, with ${p}<0.001$ between AM and AG ($2.78\pm1.80\degree$ vs $1.09\pm0.88\degree$). The orientation error $E_O$ is also decreased with the robotic guidance in both screen-based and AR-based visualization modalities compared to manual insertion. Also, the significant difference between the SG and AG ($1.47\pm1.36\degree$ vs $1.09\pm0.88\degree$) with ${p}<0.05$ illustrates the performance improvements with the AR interface.

Afterwards, the right panel of Fig.\ref{Fig_Error} shows the results of the total time cost $T_{tot}$ across all modalities, as shown in Table \ref{table:error}, the minimum value of Median and SD $76\pm55$ s is achieved in the group of AM. A significant difference of $T_{tot}$ in SM and SG ($231\pm50$ s vs $154\pm82$ s) with ${p}<0.01$ is observed, which indicates improvements in task execution. Besides, the significant difference between SM and AM ($231\pm50$ s vs $76\pm55$ s), with ${p}<0.01$, illustrates the benefits of the AR visualization interface for needle tip alignment compared to the 2D screen. It's worth noting that Fig. \ref{Fig_Error} demonstrates that the AG approach yields superior performance in terms of translation and orientation errors; while one might anticipate that the use of the integrated AG approach would result in reduced time required to complete the task, the experimental results indicate that the time cost $T_{tot}$ in AG is significantly higher than AM group ($147.5\pm109$ s vs $76\pm55$ s). This can be mainly attributed to the increased time required in the AG group to align the real needle tip with the pre-planned path (as shown at stage-2 in Fig.\ref{Fig_System}). Users reported that in the AG setup, they felt unable to make adjustments to the needle after it stiffened, whereas, in the AM setup, they still could make minor adjustments to the needle tip pose. Also, the time cost $T_{tot}$ is higher for SG than AG in Fig.\ref{Fig_Error}, as participants tend to prioritize achieving perfect alignment before performing insertion tasks with robot guidance.

Furthermore, Fig. \ref{Fig_Insetion_Error} illustrates the translation errors $E_T$ during needle insertion procedures for the four experimental setups, all conducted by the same participant. The findings indicate that procedures guided by robots exhibit greater consistency and smaller errors when compared to manually performed procedures. More specifically, in Fig.\ref{Fig_Insetion_Error} (a) and (c), it can be observed that while the needle is initially aligned well with the pre-planned path, achieving precise manipulation and executing the task accurately proves to be challenging. 

\subsection{NASA-TLX Evaluation Results}
\begin{figure*}
    \centering
    \includegraphics[scale=0.48]{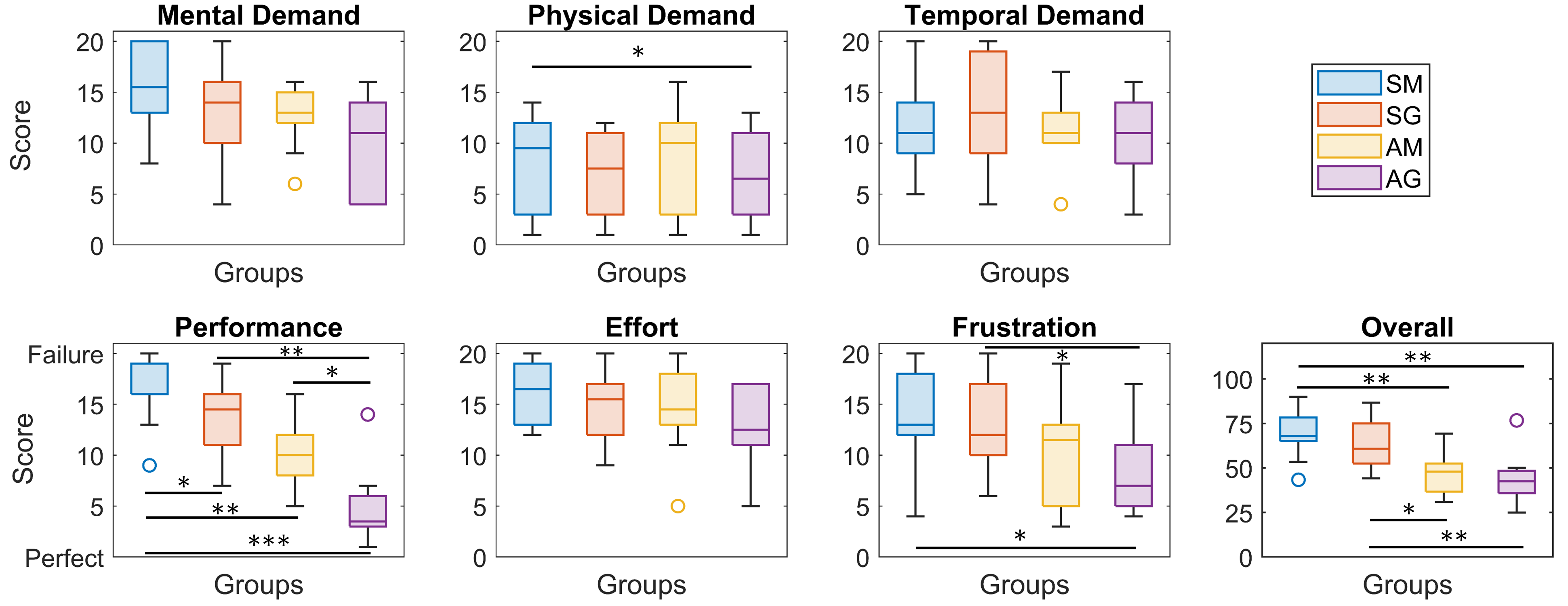}
    \caption{NASA-TLX user evaluation results from all the experimental groups. The 6 sub-items of the NASA-TLX are evaluated on a scale of the score from 0 to 20. The overall score is the sum of all the sub-items and is scaled to a range of 0 to 100. ($ *, p<0.05 ; * *, p<0.01 ; * * *, p<0.001 $)}
    \label{Fig_NASA}
\end{figure*}

The perceived workload of NASA-TLX results during the task execution is shown in Fig.\ref{Fig_NASA}, in which the last panel gives the overall workload score as well as the statistical analysis results. The proposed AG experimental setup received the lowest score among all the groups of setup with the minimum value of Median and SD $42.5\pm13.7$. The significant differences between the SM and AM ($67.9\pm13.8$ vs $47.9\pm10.8$), with ${p}<0.01$, and SG and AG ($60.8\pm14.0$ vs $42.5\pm13.7$), with ${p}<0.01$, are observed, which indicate that the AR-based visualization and navigation interface requires a lower workload compared to the 2D screen-based ones no matter with and without robotic guidance. Moreover, a significant difference, with ${p}<0.01$, between the proposed SM and the AG frameworks ($67.9\pm13.8$ vs $42.5\pm13.7$) illustrates the superiority of the proposed framework by combining the AR visualization and robotic guidance compared with traditional screen visualization and manual task execution setup. 

Specifically, although the minimum Median values from the AG group are achieved in the MD, TD, and EF items of NASA-TLX questionnaire results (11.0, 6.5, and 11.0, respectively) in Fig.\ref{Fig_Error}, no significant differences exist across all four groups of the experimental setup. Regarding the PD item, a lower score of the results and a significant difference, ${p}<0.05$, is observed with the AG framework when compared to the SM group ($9.5\pm4.8$ vs $6.5\pm4.4$). In the PE item, the participants consider that better task completion performance is achieved with SG when compared to SM ($14.5\pm3.8$ vs $19.0\pm3.6$, with ${p}<0.05$), and AM compared to AG ($10.0\pm3.8$ vs $3.5\pm3.7$, with ${p}<0.05$), which illustrates that the robotic guidance helps improve the performance. Moreover, significant differences exist when comparing the PE item score between SM and AM ($19.0\pm3.6$ vs $10.0\pm3.8$, with ${p}<0.01$), SG and AG ($14.5\pm3.8$ vs $3.5\pm3.7$, with ${p}<0.01$), which depicts that the participants believe that improved task performance has been attained with AR visualization when compared to 2D screen interface. Finally, the AG group setup achieves better FR performance when compared to SM and SG groups, respectively, with $7.0\pm4.8$ vs $13.0\pm4.9$ in AG vs SM, and $7.0\pm4.8$ vs $12.0\pm4.8$ in AG vs SG.

\subsection{Discussion}

As illustrated in the comparative analysis, the proposed framework archives better PCNL task completion performance while requiring a lower workload. This superior performance owes much to the intuitive AR visualization interface and the inclusion of robotic assistance. For instance, when we contrast our results with those of a comparable preoperative PCNL framework introduced in\cite{wang2022application}, which solely relied on the HoloLens 2 headset for visual guidance and human operator manually performed needle insertion, it becomes evident that our framework yields more consistent outcomes in terms of target kidney stone positioning accuracy (average deviation of $3.1\pm2.9$ mm in \cite{wang2022application}). Our preliminary experimental findings strongly indicate that our proposed framework holds significant promise in enhancing the efficiency and performance of PCNL task completion.

Nevertheless, it is essential to acknowledge certain limitations inherent to the proposed framework and emphasize the necessity for comprehensive studies to be undertaken. (i) Firstly, the integration of the optical tracking system and OST-HMD devices adds complexity to the system, especially considering the cluttering and crowded scenario in operating rooms (OR). In our future work, we will investigate the feasibility of leveraging the HoloLens 2 infrared (IR) camera to streamline the registration procedure, thereby simplifying the entire registration process\cite{martin2023sttar,iqbal2022semi}. Furthermore, while incorporating QR markers within the operating room might not always be feasible, an alternative approach is to attach fiducial markers to the patient's body during preoperative CT/MRI procedures\cite{lei2019mixed}. Subsequently, the geometric relationship between the target kidney stones and these fiducial markers can be obtained through anatomical structure segmentation and 3D reconstruction. (ii) Moreover, the instability issue of the HoloLens 2 OST-HMD makes it challenging to achieve a satisfactory visualization. This instability primarily stems from the self-locating capability of the OST-HMD and becomes notably problematic when participants make rapid or significant head movements. One promising solution for addressing this issue is to develop a method for estimating and compensating for the localization errors of the OST-HMD headset in three-dimensional space, thereby enhancing the accuracy of positioning\cite{lin2022robotic}. (iii) Additionally, certain critical factors that cause anatomical structure deformation, such as intricate tool-tissue interactions, patient movement, and respiration during intraoperative clinical PCNL task execution, have not been comprehensively explored. To address the challenge of compensating for kidney stone displacement relative to its preoperative position and to enhance the safety of PCNL procedures, the ultrasound imaging system can be employed. By integrating real-time ultrasound imaging into the AR interface, real-time visualization can be provided to the operators, thus, facilitating timely adjustments to the target kidney stone position and thereby improving the overall safety and precision of the procedure. (iv) Eventually, the usability verification of the proposed framework remains limited because of the experimental protocol, and non-clinical individuals were involved in the user evaluation experiment. The content, procedures, and evaluation metrics of the proposed framework should be further refined and improved to fit with the procedures in the OR. For example, more standardized metrics for evaluating the registration performance can be investigated\cite{fitzpatrick2001distribution}. Moreover, a more representative group of clinicians should be involved in the experiments to ensure a more comprehensive assessment of its usability and efficacy.

\section{Conclusion}
\label{Section: Conclusion}

This paper presents an AR and Human-Robot Collaboration control framework for robot-assisted percutaneous nephrolithotomy surgical procedures. The AR interface, with an OST-HMD employed, is adopted to intuitively display the patient hologram model, to realize interactively needle insertion path planning, as well as visual guidance during needle insertion task execution. Robotic assistance enhances the human operator's task completion performance by regulating the operator’s movement and only allowing movement along the pre-planned path. Experimental results demonstrate the superiority of the integration of AR and robotic assistance when compared to traditional 2D screen-based visualization interfaces and manual needle insertion-based task execution setups. The translation, orientation errors, and time costs are improved with the proposed framework, and a lower workload is required from the operator for the PCNL task completion.

\section*{Acknowledgments}
The authors would like to express their gratitude to all the participants in the experiment. Also, we appreciate the contributions and constructive suggestions from the Editor-in-Chief, Associate Editor, and anonymous reviewers for helping us improve this manuscript. 


\balance
\bibliographystyle{IEEEtran}
\bibliography{Main}

\begin{thebibliography}{10}
\providecommand{\url}[1]{#1}
\csname url@samestyle\endcsname
\providecommand{\newblock}{\relax}
\providecommand{\bibinfo}[2]{#2}
\providecommand{\BIBentrySTDinterwordspacing}{\spaceskip=0pt\relax}
\providecommand{\BIBentryALTinterwordstretchfactor}{4}
\providecommand{\BIBentryALTinterwordspacing}{\spaceskip=\fontdimen2\font plus
\BIBentryALTinterwordstretchfactor\fontdimen3\font minus \fontdimen4\font\relax}
\providecommand{\BIBforeignlanguage}[2]{{%
\expandafter\ifx\csname l@#1\endcsname\relax
\typeout{** WARNING: IEEEtran.bst: No hyphenation pattern has been}%
\typeout{** loaded for the language `#1'. Using the pattern for}%
\typeout{** the default language instead.}%
\else
\language=\csname l@#1\endcsname
\fi
#2}}
\providecommand{\BIBdecl}{\relax}
\BIBdecl

\bibitem{de2015percutaneous}
S.~De, R.~Autorino, F.~J. Kim \emph{et~al.}, ``Percutaneous nephrolithotomy versus retrograde intrarenal surgery: a systematic review and meta-analysis,'' \emph{European urology}, vol.~67, no.~1, pp. 125--137, 2015.

\bibitem{rodrigues2013kidney}
P.~L. Rodrigues, N.~F. Rodrigues, J.~Fonseca, E.~Lima, and J.~L. Vila{\c{c}}a, ``Kidney targeting and puncturing during percutaneous nephrolithotomy: recent advances and future perspectives,'' \emph{Journal of endourology}, vol.~27, no.~7, pp. 826--834, 2013.

\bibitem{jiang2021autonomous}
Z.~Jiang, Z.~Li, M.~Grimm, M.~Zhou, M.~Esposito \emph{et~al.}, ``Autonomous robotic screening of tubular structures based only on real-time ultrasound imaging feedback,'' \emph{IEEE Transactions on Industrial Electronics}, vol.~69, no.~7, pp. 7064--7075, 2021.

\bibitem{paranawithana2018ultrasound}
I.~Paranawithana, H.-Y. Li, S.~Foong, U.-X. Tan, L.~Yang, T.~S.~K. Lim, and F.~C. Ng, ``Ultrasound-guided involuntary motion compensation of kidney stones in percutaneous nephrolithotomy surgery,'' in \emph{2018 IEEE 14th International Conference on Automation Science and Engineering (CASE)}, pp. 1123--1129.\hskip 1em plus 0.5em minus 0.4em\relax IEEE, 2018.

\bibitem{li2018towards}
H.-Y. Li, I.~Paranawithana, Z.~H. Chau, L.~Yang \emph{et~al.}, ``Towards to a robotic assisted system for percutaneous nephrolithotomy,'' in \emph{2018 IEEE/RSJ International Conference on Intelligent Robots and Systems (IROS)}, pp. 791--797.\hskip 1em plus 0.5em minus 0.4em\relax IEEE, 2018.

\bibitem{tu2022ultrasound}
P.~Tu, C.~Qin, Y.~Guo, D.~Li, A.~J. Lungu, H.~Wang, and X.~Chen, ``Ultrasound image guided and mixed reality-based surgical system with real-time soft tissue deformation computing for robotic cervical pedicle screw placement,'' \emph{IEEE Transactions on Biomedical Engineering}, vol.~69, no.~8, pp. 2593--2603, 2022.

\bibitem{stoianovici1998modular}
D.~Stoianovici, L.~L. Whitcomb, J.~H. Anderson \emph{et~al.}, ``A modular surgical robotic system for image guided percutaneous procedures,'' in \emph{Medical Image Computing and Computer-Assisted Intervention—MICCAI’98: First International Conference Cambridge, MA, USA, October 11--13, 1998 Proceedings 1}, pp. 404--410.\hskip 1em plus 0.5em minus 0.4em\relax Springer, 1998.

\bibitem{ferraguti2022augmented}
F.~Ferraguti, S.~Farsoni, and M.~Bonf{\`e}, ``Augmented reality and robotic systems for assistance in percutaneous nephrolithotomy procedures: Recent advances and future perspectives,'' \emph{Electronics}, vol.~11, no.~19, p. 2984, 2022.

\bibitem{wilz2021constrained}
O.~Wilz, B.~Sainsbury, and C.~Rossa, ``Constrained haptic-guided shared control for collaborative human--robot percutaneous nephrolithotomy training,'' \emph{Mechatronics}, vol.~75, p. 102528, 2021.

\bibitem{sainsbury2022preoperative}
B.~Sainsbury, O.~Wilz, J.~Ren, M.~Green, M.~Fergie, and C.~Rossa, ``Preoperative virtual reality surgical rehearsal of renal access during percutaneous nephrolithotomy: A pilot study,'' \emph{Electronics}, vol.~11, no.~10, p. 1562, 2022.

\bibitem{farcas2021simulation}
M.~Farcas, L.~F. Reynolds, and J.~Y. Lee, ``Simulation-based percutaneous renal access training: evaluating a novel 3d immersive virtual reality platform,'' \emph{Journal of Endourology}, vol.~35, no.~5, pp. 695--699, 2021.

\bibitem{fu2023recent}
J.~Fu, A.~Rota, S.~Li, J.~Zhao, Q.~Liu, E.~Iovene, G.~Ferrigno, and E.~De~Momi, ``Recent advancements in augmented reality for robotic applications: A survey,'' in \emph{Actuators}, vol.~12, no.~8, p. 323.\hskip 1em plus 0.5em minus 0.4em\relax MDPI, 2023.

\bibitem{palumbo2022mixed}
M.~C. Palumbo, S.~Saitta, M.~Schiariti, M.~C. Sbarra, E.~Turconi, G.~Raccuia, J.~Fu, V.~Dallolio, P.~Ferroli, E.~Votta \emph{et~al.}, ``Mixed reality and deep learning for external ventricular drainage placement: A fast and automatic workflow for emergency treatments,'' in \emph{International Conference on Medical Image Computing and Computer-Assisted Intervention}, pp. 147--156.\hskip 1em plus 0.5em minus 0.4em\relax Springer, 2022.

\bibitem{li2021towards}
R.~Li, Y.~Tong, T.~Yang, J.~Guo, W.~Si, Y.~Zhang, R.~Klein, and P.-A. Heng, ``Towards quantitative and intuitive percutaneous tumor puncture via augmented virtual reality,'' \emph{Computerized Medical Imaging and Graphics}, vol.~90, p. 101905, 2021.

\bibitem{wang2022application}
L.~Wang, Z.~Zhao, G.~Wang \emph{et~al.}, ``Application of a three-dimensional visualization model in intraoperative guidance of percutaneous nephrolithotomy,'' \emph{International Journal of Urology}, vol.~29, no.~8, pp. 838--844, 2022.

\bibitem{qian2019review}
L.~Qian, J.~Y. Wu, S.~P. DiMaio, N.~Navab, and P.~Kazanzides, ``A review of augmented reality in robotic-assisted surgery,'' \emph{IEEE Transactions on Medical Robotics and Bionics}, vol.~2, no.~1, pp. 1--16, 2019.

\bibitem{lin2021arei}
Z.~Lin, A.~Gao, X.~Ai, H.~Gao, Y.~Fu, W.~Chen, and G.-Z. Yang, ``Arei: Augmented-reality-assisted touchless teleoperated robot for endoluminal intervention,'' \emph{IEEE/ASME Transactions on Mechatronics}, vol.~27, no.~5, pp. 3144--3154, 2021.

\bibitem{tsai1989new}
R.~Y. Tsai, R.~K. Lenz \emph{et~al.}, ``A new technique for fully autonomous and efficient 3 d robotics hand/eye calibration,'' \emph{IEEE Transactions on robotics and automation}, vol.~5, no.~3, pp. 345--358, 1989.

\bibitem{arun1987least}
K.~S. Arun, T.~S. Huang, and S.~D. Blostein, ``Least-squares fitting of two 3-d point sets,'' \emph{IEEE Transactions on pattern analysis and machine intelligence}, no.~5, pp. 698--700, 1987.

\bibitem{fu2023augmented}
J.~Fu, M.~C. Palumbo, E.~Iovene, L.~Qingsheng, B.~Ilaria, A.~Redaelli, G.~Ferrigno, and E.~De~Momi, ``Augmented reality-assisted robot learning framework for minimally invasive surgery task,'' in \emph{2023 IEEE International Conference on Robotics and Automation (ICRA)}, pp. 11\,647--11\,653.\hskip 1em plus 0.5em minus 0.4em\relax IEEE, 2023.

\bibitem{hart2006nasa}
S.~G. Hart, ``Nasa-task load index (nasa-tlx); 20 years later,'' in \emph{Proceedings of the human factors and ergonomics society annual meeting}, vol.~50, no.~9, pp. 904--908.\hskip 1em plus 0.5em minus 0.4em\relax Sage publications Sage CA: Los Angeles, CA, 2006.

\bibitem{benmahdjoub2023evaluation}
M.~Benmahdjoub, A.~Thabit, M.-L.~C. van Veelen, W.~J. Niessen, E.~B. Wolvius, and T.~van Walsum, ``Evaluation of ar visualization approaches for catheter insertion into the ventricle cavity,'' \emph{IEEE Transactions on Visualization and Computer Graphics}, 2023.

\bibitem{martin2023sttar}
A.~Martin-Gomez, H.~Li, T.~Song, S.~Yang, G.~Wang, H.~Ding, N.~Navab, Z.~Zhao, and M.~Armand, ``Sttar: surgical tool tracking using off-the-shelf augmented reality head-mounted displays,'' \emph{IEEE Transactions on Visualization and Computer Graphics}, 2023.

\bibitem{iqbal2022semi}
H.~Iqbal and F.~R. y~Baena, ``Semi-automatic infrared calibration for augmented reality systems in surgery,'' in \emph{2022 IEEE/RSJ International Conference on Intelligent Robots and Systems (IROS)}, pp. 4957--4964.\hskip 1em plus 0.5em minus 0.4em\relax IEEE, 2022.

\bibitem{lei2019mixed}
P.-f. Lei, S.-l. Su, L.-y. Kong, C.-g. Wang, D.~Zhong, and Y.-h. Hu, ``Mixed reality combined with three-dimensional printing technology in total hip arthroplasty: An updated review with a preliminary case presentation,'' \emph{Orthopaedic Surgery}, vol.~11, no.~5, pp. 914--920, 2019.

\bibitem{lin2022robotic}
Z.~Lin, T.~Zhang, Z.~Sun, H.~Gao, X.~Ai, W.~Chen, G.-Z. Yang, and A.~Gao, ``Robotic telepresence based on augmented reality and human motion mapping for interventional medicine,'' \emph{IEEE Transactions on Medical Robotics and Bionics}, vol.~4, no.~4, pp. 935--944, 2022.

\bibitem{fitzpatrick2001distribution}
J.~M. Fitzpatrick and J.~B. West, ``The distribution of target registration error in rigid-body point-based registration,'' \emph{IEEE transactions on medical imaging}, vol.~20, no.~9, pp. 917--927, 2001.

\end{thebibliography}

\end{document}